# Fully Convolutional Change Detection Framework with Generative Adversarial Network for Unsupervised, Weakly Supervised and Regional Supervised Change Detection

Chen Wu, *Member*, *IEEE*, Bo Du, *Senior Member*, *IEEE*, Liangpei Zhang, *Fellow*, *IEEE*

**Abstract**—Deep learning for change detection is one of the current hot topics in the field of remote sensing. However, most end-to-end networks are proposed for supervised change detection, and unsupervised change detection models depend on traditional pre-detection methods. Therefore, we proposed a fully convolutional change detection framework with generative adversarial network, to conclude unsupervised, weakly supervised, regional supervised, and fully supervised change detection tasks into one framework. A basic Unet segmentor is used to obtain change detection map, an image-to-image generator is implemented to model the spectral and spatial variation between multi-temporal images, and a discriminator for changed and unchanged is proposed for modeling the semantic changes in weakly and regional supervised change detection task. The iterative optimization of segmentor and generator can build an end-to-end network for unsupervised change detection, the adversarial process between segmentor and discriminator can provide the solutions for weakly and regional supervised change detection, the segmentor itself can be trained for fully supervised task. The experiments indicate the effectiveness of the propsed framework in unsupervised, weakly supervised and regional supervised change detection. This paper provides theorical definitions for unsupervised, weakly supervised and regional supervised change detection tasks, and shows great potentials in exploring end-to-end network for remote sensing change detection.

**Index Terms**— Remote Sensing, Change Detection, Fully Covolutional Network, Generative Adversarial Network, Weakly Supervised Segmentation

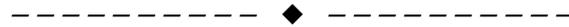

——————————— ◆ ———————————

# 1 Introduction

Change detection is one of the earliest and most widely used technology in the field of remote sensing [1-6]. It aims at finding landscape changes from the multi-temporal remote sensing images observing the same study site at different time. It has been widely used in land-use/land-cover change analysis [7, 8], urban study and environmental monitoring [9, 10], and disaster assessment [11].

There are lots of existing researches about how to extract and label changes more accurately and effectively [12-16]. With the rising of deep learning in computer vision, it can also be applied in remote sensing change detection in recent years [17]. The deep learning networks were firstly used as feature extractor embedded in the traditional process of change detection [18-23]. The minization of feature difference of unchanged patches is always selected as the learning criterion [20, 24]. Another way for deep learning-based unsupervised change detection, is to distinguish certain changes and non-changes by pre-detection, and use the corresponding patches as training samples to build a deep network model to extract better features and discriminate semantic labels [25-27]. However, these kinds of methods deeply depend on pre-detection result, and cannot build an advanced end-to-end model to make full use of deep learning.

With the sharing of labelled change detection dataset, the supervised semantic segmentation was brought into the field of change detection, for identifying binary changes or targeted changes from high resolution images with fully convolutional networks [28-30]. The abundant datasets make the change detection models obtain higher and higher accuracies [31-33]. Even though these works utilized the end-to-end fully segmentation change detection model and show satisfactory performances, it is must be noting that labelling all the changed pixels in an image as the training samples of one dataset is quite time and labor consuming, and needs more professional knowledge compared with the tasks in computer vision.

With the fixation of fully convolutional segmentation network for processing multi-temporal high-resolution remote sensing images, change detection task can be defined as obtaining a suitable segmentation map to satisfy the pre-

- *C. Wu is with the State Key Laboratory of Information Engineering in Surveying, Mapping and Remote Sensing, Wuhan University, Wuhan 430072, China. E-mail: chen.wu@whu.edu.cn.*
- *B. Du is with National Engineering Research Center for Multimedia Software, Institute of Artificial Intelligence, School of Computer Science and Hubei Key Laboratory of Multimedia and Network Communication Engineering, Wuhan University, Wuhan, China. E-mail: gunspace@163.com.*
- *L. Zhang is with the State Key Laboratory of Information Engineering in Surveying, Mapping and Remote Sensing, Wuhan University, Wuhan 430072, China. E-mail: zlp62@whu.edu.cn.*

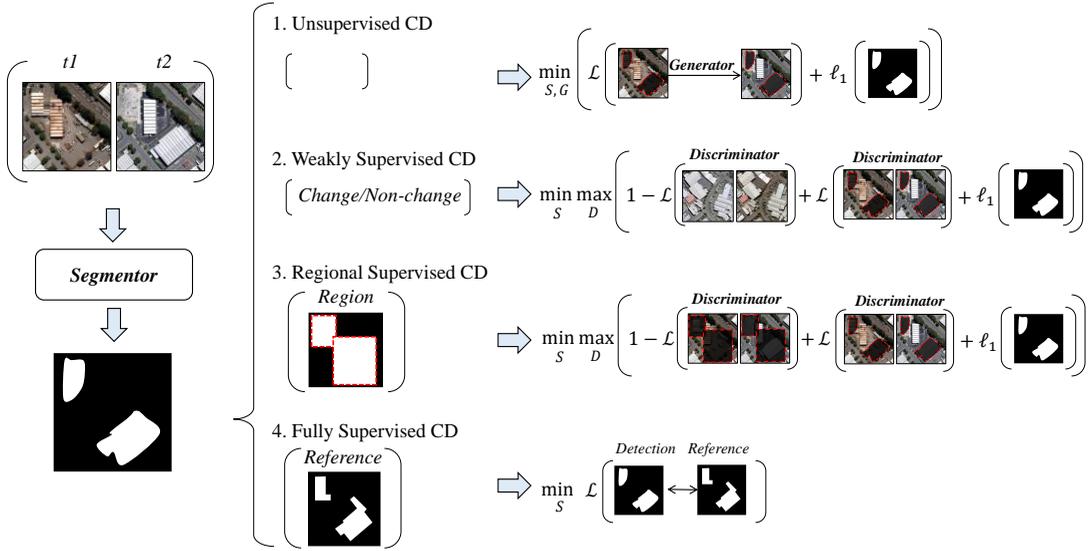

Figure 1 The framework of Fully Convolutional Change Detection with Generative Adversarial Network.

defined constraints. The task of *fully supervised change detection (FSCD)* can be considered as the full constraint to the reference map [34, 35]. However, it is hard to always provide abundant FSCD training dataset in practical applications, since change detection is a complex task and needs to interpret multi-temporal image pair at the same time. So, can we release the constraint to build an end-to-end network framework for satisfying the practical needs?

For *unsupervised change detection (USCD)*, although we have no pre-given samples or rules, there is a prior knowledge that, the representation of unchanged landscapes can be predicated from one image to another image, while the changed landscape cannot. So, we can make a constraint for unsupervised change detection, that the model wants to find a segmentation region, without which the transitions between multi-temporal images can be better predicted.

Furthermore, we can also define *weakly supervised change detection (WSCD)* task, that only the label of change or non-change is given to a pair of multi-temporal images, and we want to extract the exact changed regions. The constraint is, the model aims at finding a segmentation region, without which the image pairs containing changed landscape can be determined as unchanged pairs.

Finally, we will define a new change detection task, that we only need to draw a probable region (such as a rectangle) of landscape changes, and the exact changed pixels are labelled accurately. The constraint is, the model labels a change region, so that the image pair with the mask of segmentation map can be determined as unchanged just like that with the mask of pre-defined region. We call it *regional supervised change detection (RSCD)*.

In order to conclude all these change detection tasks, in this paper, we propose a framework named fully convolutional change detection with generative adversarial network (FCD-GAN) for unsupervised, weakly supervised and regional supervised change detection task. A basic segmentor for multi-temporal high-resolution remote sensing images was used to obtain change maps with various constraints for different tasks. The unsupervised change detection model is built with an iterative optimization of segmentor and a generator, that predicts one image to the corresponding multi-temporal image. The weakly supervised model was optimized with the adversarial balance between segmentor to find more accurate changed regions and discriminator to better distinguish changed and unchanged pairs. The regional supervised model aims to train a segmentor to mask the change region so that this image pair can be determined as unchanged with the discriminator.

The contributions of this paper can be listed as follows:

1) We proposed a framework named fully convolutional change detection with generative adversarial network (FCD-GAN), which can conclude unsupervised, weakly supervised, regional supervised, and fully supervised change detection into one framework.

2) This is the first time that unsupervised change detection can be achieved by fully convolutional network, which provides significant potentials for the further development of unsupervised change detection.

3) As far as we know, few studies utilize GAN structure in one-stage weakly supervised segmentation. For the first time, this paper implemented the task of weakly supervised change detection, and show a new approach for weakly supervised segmentation with GAN.

4) For the first time, this paper proposed a task of regional supervised change detection, which is very useful in practical applications. This task is achieved also by GAN constraints in the proposed framework.

## 2 METHODOLOGY

The proposed framework is shown in Figure 1. It consists of three basic modules: segmentor, generator and discriminator. Since the focus of this paper is to propose the common framework, we only use very simple network structures in these three modules. In the task of unsupervised, weakly supervised and regional supervised change detection, the $\ell_1$-norm constraint of segmentation map is necessary to avoid all-change output. This norm will be detailed introduced in section 2.2. Since the fully supervised change detection with the pixel-level training references is very common, we don't focus on this task in our paper. The details of implementation are as follows:

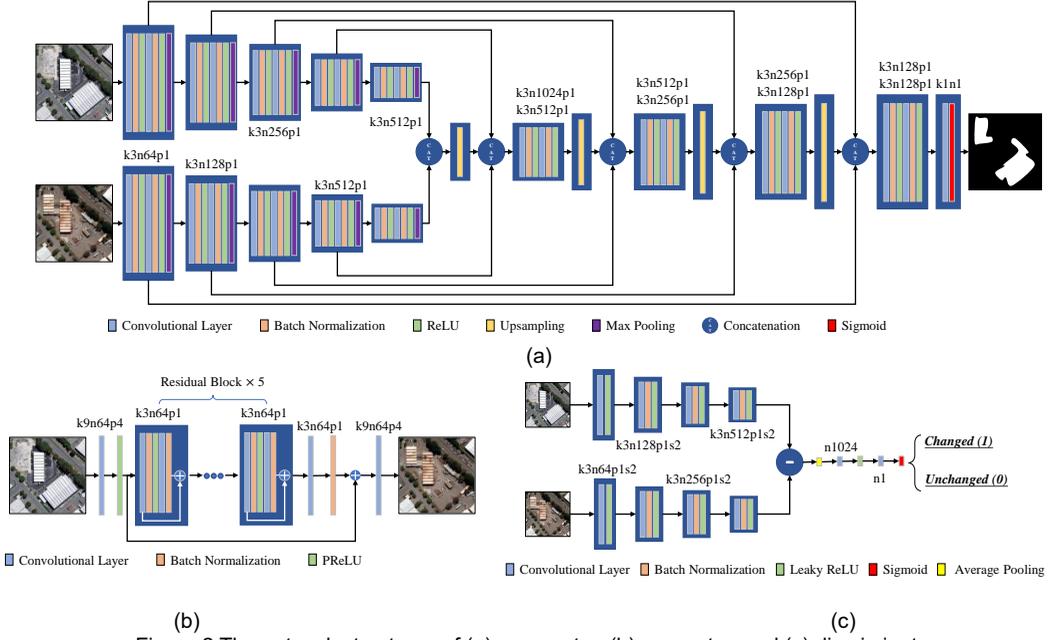

Figure 2 The network structures of (a) segmentor, (b) generator, and (c) discriminator.

### 2.1 Basic Module

In current studies, lots of advanced segmentation models and tricks have been used in change detection [31, 33, 36-38]. However, since the focus of this paper is only to propose a framework, we only use a very basic structure for segmentor, generator, and discriminator.

For segmentor, we choose a basic U-net structure, which has been used in many change detection researches [28, 30]. Since multi-temporal high-resolution images should be fused in encoder, we selected a simple concatenate fusion in every layer. Sigmoid is used as the final activation function to ensure the range of output to be from 0 to 1. The two branches of encoder are Siamese. The upsampling approach is bilinear interpolation. The structure of segmentor is shown in Figure 2 (a).

For generator, it can be assumed that the multi-temporal images can be predicted from one image to another if the landscapes don't change. Therefore, we implement a generator just like that from SRGAN [39]. The difference is that no activation function such as sigmoid or tanh is used in the last layer, since the input multi-temporal images will be pre-processed by standard normalization. The structure of generator is shown in Figure 2 (b).

For discriminator, we simplify the discrimination network in SRGAN [39]. In the research, we found that complex network will be too powerful in distinguishing changed or unchanged image scene, so that the balance in adversarial progress is too easy to break. The branches of encoder are Siamese. So, the simplified discriminator is shown in Figure 2 (c).

### 2.2 Unsupervised Change Detection

In most previous unsupervised change detection studies with deep learning models, the networks were trained with pre-detection results and small patches centering the detected pixels [22, 29, 40-42]. In this paper, we want to utilize an end-to-end fully convolutional network to deal with this task.

The basic assumption of unsupervised change detection is: for the unchanged landscapes in multi-temporal images, they have some consistency in spectral, spatial or semantic aspects; while for the changed landscapes, they can be any temporal variations [3]. So, the task can be summarized as:

*The model wants to find a change region, so that in the remained regions one image can be predicted into another multi-temporal image accurately.*

In theory, $\mathbf{X}$ and $\mathbf{Y}$ are the multi-temporal images, $g(\cdot)$ is the generator, and $s(\cdot)$ is the segmentator. The objective can be written as:

$$\min_{g,s}\left[\mathcal{L}\left(g(\mathbf{X})\cdot(1-s(\mathbf{X},\mathbf{Y})), \mathbf{Y}\cdot(1-s(\mathbf{X},\mathbf{Y}))\right)\right] \quad (1)$$

where $\mathcal{L}(\cdot)$ is the loss function to measure the similarity between the predicted image and the real image, and $s(\mathbf{X},\mathbf{Y})$ will segment every pixel in the image to be non-change (0) or change (1).

However, it worth noting that there is one way to satisfy this objective easily, which is labelling all the pixels to be 1. Obviously, it is a false solution. In order to avoid this meaningless solution, we need to add a constraint that the segmentation should be 0/1 and also sparse. The most straightforward way is to add a 0-1 loss in the loss function, whereas $\ell_0$ loss is hard to optimize. So, we release this constraint to $\ell_1$ loss. Therefore, the objective is rewritten as:

$$\min_{g,s}[\mathcal{L}_g + \lambda\ell_1(s(\mathbf{X},\mathbf{Y}))] \quad (2)$$

where $\lambda$ indicates the weight of $\ell_1$ loss, and $\mathcal{L}_g$ indicates the loss function of generator with the change mask. Mostly, higher weight on $\ell_1$ loss will lead to fewer segmented changes with higher precision rate and lower recall rate.

In this way, the final task of unsupervised change detection is summarized as:

*The model wants to find a <u>minimum</u> change region, so that in the remained regions one image can be predicted into another multi-temporal image accurately.*

The loss function $\mathcal{L}_g$ to measure the prediction in this paper is a combination of reconstruction loss and content loss. For the reconstruction loss, since it is affected by the segmentation output, we use a weighted $\ell_1$ loss to measure the mean prediction error; for the content loss, we choose the loss function as SRGAN [39], where the MSE loss of the 29$^{th}$ layer features from the pretrained vgg16 network is used to calculate the content difference. The generation loss function can be written as:

$$\mathcal{L}_g = \frac{\sum(1-s(\mathbf{X},\mathbf{Y})_{ij})|g(\mathbf{X})_{ij}-\mathbf{Y}_{ij}|}{\sum(1-s(\mathbf{X},\mathbf{Y})_{ij})} + \mu \ell_{MSE}\left(\phi_{vgg}^{29}(g(\mathbf{X})\cdot(1-s(\mathbf{X},\mathbf{Y})_{ij})),\phi_{vgg}^{29}(\mathbf{Y}\cdot(1-s(\mathbf{X},\mathbf{Y})_{ij}))\right) \quad (3)$$

where $\phi_{vgg}^{29}(\cdot)$ indicates the output of the 29$^{th}$ layer features from vgg16 network in pytorch, and $\mu$ is the weight of content loss. It can be found that multi-level convolutional features can be summed up in vgg16 network. In this paper, we only use the 29$^{th}$ layer features. Another thing needing attention is that the pretrained vgg16 network only has three channels for its input, whereas remote sensing imagery always contains four (R, G, B and NIR) or more bands. So, the loss band by band was calculated to obtain the mean value.

Since there are two network models needing learning and training, they are optimized with iteration progress. The generator was firstly trained with the whole image. Then, the segmentor was trained with the pre-trained generator. And finally, the generator and segmentor are trained together, iteratively. Adam optimizer and warm-up strategy are used in optimization.

## 2.3 Weakly Supervised Change Detection

In change detection, it is hard and time-consuming to label all the changed pixels from the multi-temporal images. So, if we can detect changes accurately only by labelling the image pair to be changed (1) or unchanged (0), it is more convenient in practical applications. It is also feasible to define the target changes in semantic aspect.

For weakly supervised change detection, the interpreters label numerous multi-temporal image pairs to be changed or unchanged. We want to train a segmentation model with the weakly supervised information to detect exact changed pixels in these change image pairs.

The basic assumption for solving this problem is that, if the changed pixels are all labelled and masked, the corresponding changed image pairs will be identified as the same as the unchanged pairs. It can be found that this is an adversarial process, as shown in Figure 3. With the advance of segmentor, it is harder and harder for the developing discriminator to distinguish the masked changed pairs and the unchanged pairs.

In this way, the final task of weakly supervised change

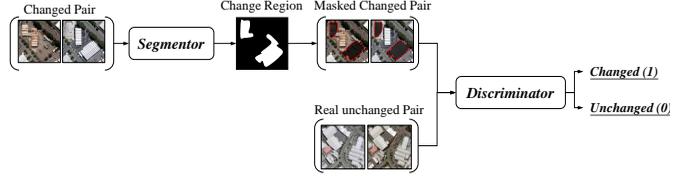

Figure 3 The adversarial process of weakly supervised change detection.

detection is summarized as:

*The model wants to find a minimum change region, so that without the changed regions the changed image pairs can be discriminated as the same as the real unchanged image pairs.*

When the loss function of the proposed model is designed, we utilize the format like WGAN for better convergence [43]. Also, the generation loss can be used for better optimization. The generator can be trained only with the real unchanged image pairs, and the generation loss can help to find the change regions in the changed pairs. However, it is worth noting that when weakly supervised information is given, maybe not all the unpredictable regions are target change regions, such as phenological changes. So, the generation loss must be adjusted with a weight. Besides, when the segmentor is applied on the real unchanged image pairs, the results should be absolutely zero throughout the whole image. Thus, we use a $\ell_2$ loss to take advantage of this constraint. Finally, the objective of the weakly supervised change detection can be written as:

$$\min_d [1 - \mathcal{L}_d^c + \mathcal{L}_d^u] \quad (4)$$

$$\min_s [\mathcal{L}_d^c + \lambda_1 \ell_1(s(\mathbf{X}^c,\mathbf{Y}^c)) + \lambda_2 \ell_2(s(\mathbf{X}^u,\mathbf{Y}^u)) + \lambda_3 \mathcal{L}_g] \quad (5)$$

where $\mathcal{L}_d^u$ and $\mathcal{L}_d^c$ indicate the discrimination loss for real unchanged pairs and changed pairs, $\mathbf{X}^c$, $\mathbf{Y}^c$ indicate the changed multi-temporal image pair, $\mathbf{X}^u$, $\mathbf{Y}^u$ indicate the real unchanged multi-temporal image pair, $\mathcal{L}_g$ indicates the generation loss, a series of $\lambda$ indicates the weights of different losses. $\mathcal{L}_g$ can be calculated by ( 3 ) only for changed image pairs.

For detailed discrimination loss functions, they can be written as:

$$\mathcal{L}_d^u = d\left(\mathbf{X}^u \cdot (1-s(\mathbf{X}^c,\mathbf{Y}^c)), \mathbf{Y}^u \cdot (1-s(\mathbf{X}^c,\mathbf{Y}^c))\right) \quad (6)$$

$$\mathcal{L}_d^c = d\left(\mathbf{X}^c \cdot (1-s(\mathbf{X}^c,\mathbf{Y}^c)), \mathbf{Y}^c \cdot (1-s(\mathbf{X}^c,\mathbf{Y}^c))\right) \quad (7)$$

It worth noting that the discrimination loss of unchanged image pair is also calculated with a change mask. If not, the discriminator will converge to zero quickly, and the balance in adversarial process will be easily broken. It is because, the discriminator tends to learn whether there is a mask covering the image pair instead of they are a changed pair or not. We all know that: an unchanged multi-temporal image pair will always keep unchanged with any mask. Therefore, the changed mask will also be applied on both the changed image pair and real unchanged image pair.

Due to WGAN [43], RMSProp optimizer is used for the adversarial process. The learning rate of discriminator is determined much lower than that of segmentor, for better optimization.

## 2.4 Regional Supervised Change Detection

In practical applications, even though labeling all changed pixels is labor- and time-consuming, it is also too coarse to just provide an image-level label for weakly supervised change detection. Therefore, we propose a new task named regional supervised change detection (RSCD).

In regional supervised change detection, the interpreters only draw a probable region, such as a rectangle, covering the landscape changes. The model in this task needs to find the extract changed pixels inside the labelled region.

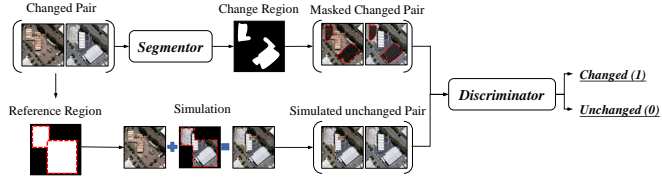

Figure 4 The adversarial process of regional supervised change detection.

The basic assumption is, if the changed pixels are all masked, the image pair can be seen as an unchanged one just like that by masking the given reference region. Similar with weakly supervised change detection, this task can be solved with an adversarial process. The difference is the unchanged pair sample is a simulated one by masking the pre-given reference region, instead of a real one. In implementation, if the simulated unchanged pair is only made by simple masking, the adversarial process is hard to maintain, since the discriminator is easy to find a regular mask in the image pair. So, the simulation is achieved by replacing the content inside the reference region from one multi-temporal image to the other, as shown in Figure 4.

Therefore, the final task of regional supervised change detection is summarized as:

*The model wants to find a minimum change region, so that without the changed regions the changed image pairs can be discriminated as unchanged as the simulated unchanged image pairs by replacing the content inside the pre-given reference region.*

The objective of regional supervised change detection can be formulated derived from ( 4 ) as:

$$\min_{d}[1 - \mathcal{L}_d^c + \mathcal{L}_d^u] \quad (8)$$

$$\min_{s}[\mathcal{L}_d^c + \lambda_1 \ell_1(s(\mathbf{X},\mathbf{Y}) \cdot \mathbf{R}) + \lambda_2 \ell_2(s(\mathbf{X},\mathbf{Y}) \cdot (1 - \mathbf{R})) + \lambda_3 \mathcal{L}_g] \quad (9)$$

where $\mathbf{R}$ indicates the reference region. It means the model should tend to find sparse changed pixels inside the reference region, and no pixels outside the reference.

For the discrimination loss, they can also be written as:

$$\mathcal{L}_d^u = d\left(\mathbf{X} \cdot \big(1 - s(\mathbf{X},\mathbf{Y})\big), \widehat{\mathbf{Y}} \cdot \big(1 - s(\mathbf{X},\mathbf{Y})\big)\right) \quad (10)$$

$$\mathcal{L}_d^c = d\left(\mathbf{X} \cdot \big(1 - s(\mathbf{X},\mathbf{Y})\big), \mathbf{Y} \cdot \big(1 - s(\mathbf{X},\mathbf{Y})\big)\right) \quad (11)$$

where $\widehat{\mathbf{Y}}$ is the simulated unchanged image, which can be generated by $\widehat{\mathbf{Y}} = \mathbf{Y} \cdot (1 - \mathbf{R}) + \mathbf{X} \cdot \mathbf{R}$.

The optimization of regional supervised change detection is the same as that of weakly supervised change detection

## 3 DATASETS

For the experiment of unsupervised change detection, we choose two high-resolution multi-temporal GF-2 remote sensing image datasets, as shown in Figure 5 and Figure 6. These two high-resolution multi-temporal remote sensing image datasets were all acquired by GF-2 satellite. The time-1 image was acquired on April 4, 2016, and the time-2 image was acquired on September 1, 2016. The size of multi-temporal images is $1000 \times 1000$, with four spectral bands (Blue, Green, Red, NIR) and the spatial resolution of 4m. These two datasets were used for unsupervised change detection in the previous study [44].

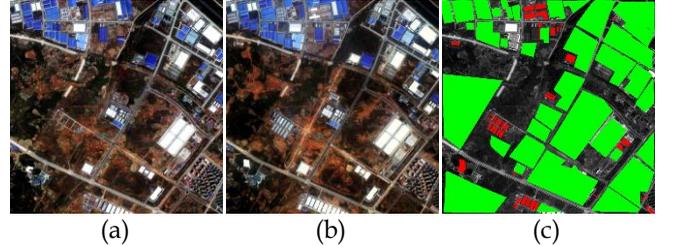

Figure 5 Wuhan multi-temporal GF-2 image dataset, (a) time-1 image, (b) time-2 image, and (c) reference, where red indicates changed pixels and green indicates unchanged pixels.

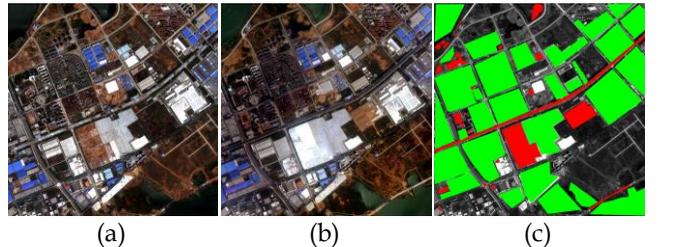

Figure 6 Hanyang multi-temporal GF-2 image dataset, (a) time-1 image, (b) time-2 image, and (c) reference, where red indicates changed pixels and green indicates unchanged pixels.

For the experiments of weakly and regional supervised change detection, since there is no corresponding dataset, we generate two datasets with shared dataset for supervised change detection.

First, we choose the WHU building change detection dataset (BCD) [45]. This dataset contains two aerial high-resolution images, which were acquired in 2012 and 2016. The size of study image is $32507 \times 15354$, with 3 optical bands and the resolution of 0.3m. Building change is the target change of this dataset. In order to generate a weakly supervised change detection dataset, we slice these two large images into $200 \times 200$ small images, as well as the change reference. If there are any changes in the image scene, this image pair will be labelled as changed. So finally, we got 9935 unchanged pairs and 2616 changed pairs, as shown in Figure 7. It can be found that this weakly supervised change detection dataset is very complex and challenging, since some landscape changes are not defined as target semantic changes.

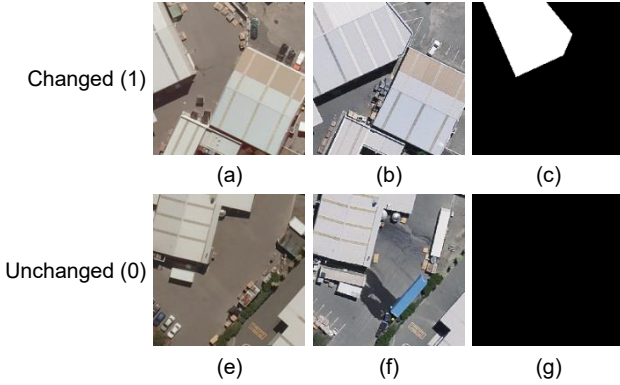

Figure 7 Samples for weakly supervised change detection, including a changed pair (a) before change, (b) after change, (c) reference, and an unchanged pair (c) before change, (b) after change, (c) reference.

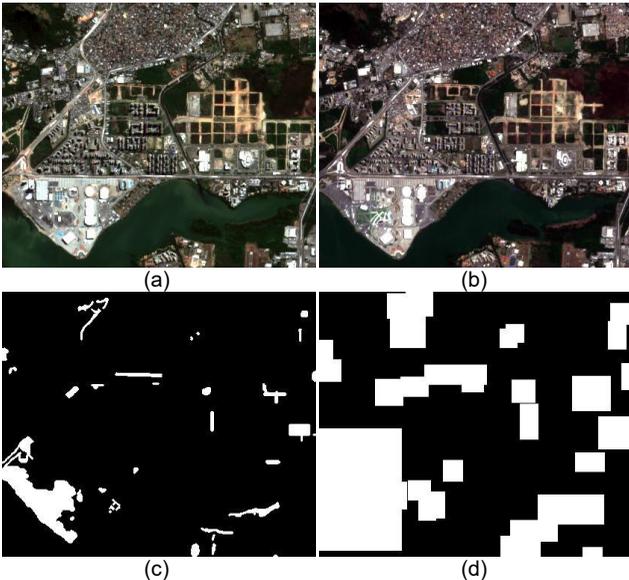

Figure 8 A multi-temporal image pair covering Rio which are acquired on (a) April 24, 2016, (b) October 11, 2017, (c) pixel-level reference, and (d) region-level reference.

Then, we choose Onera Satellite Change Detection dataset (OSCD) to evaluate the regional supervised change detection task [46]. This dataset comprised 24 pairs of multi-temporal images, acquired by Sentinel-2 satellite. It contains 14 training and 10 test image pairs. The remote sensing imagery from Sentinel-2 satellite has 13 spectral bands with the spatial resolution of 10, 20m and 60. For flexible process, we only choose four spectral bands (B, G, R and NIR) with 10m resolution. Co-registration was implemented to reduce the mis-registration errors in some pairs, and common subsets were selected to keep the same size of multi-temporal image pair and reference. In the generation of RSCD, the connected regions of change refence are all extracted, and a bounding box was drawn. In order to simulate the manual interpretation, an expansion of 10 pixels was applied to the bounding boxes in four directions. In this way, we finally obtain a dataset with pixel-based references and region-based supervision. The example of multi-temporal image pair and its references are shown in Figure 8. In this task, we only have the region-level reference, and we want to train a segmentation model to find the extract pixel-level change detection result. Since OSCD has distinguished training and testing samples, the RSCD model is also trained on the training pairs and evaluated on the testing pairs.

## 4 EXPERIMENTS

### 4.1 Unsupervised Change Detection

Since the proposed model is a fully convolutional model, it is hard to process a whole image with the size of 1000 × 1000 directly. So, the large image will be sliced into small patches for batch process. In order to avoid the problem in processing patch edge, the patches are sliced with overlapping. In USCD experiment, the patch size is 220 × 220 with an overlapping of 10 pixels in each direction. Only the output of the centered 200 × 200 patch will be used in the final result.

TABLE 1 ACCURACY ASSESSMENT ON WH DATASET

|  | OA | KC | Pre. | Rec. | F1 | mIOU | cIOU |
|---|---|---|---|---|---|---|---|
| **RNN** | 0.9747 | 0.6517 | 0.7006 | 0.6325 | 0.6648 | 0.7360 | 0.4979 |
| **DSCN** | 0.9732 | 0.5582 | 0.7855 | 0.4484 | 0.5709 | 0.6861 | 0.3995 |
| **SiamCRNN_FC** | 0.9767 | 0.6919 | 0.7098 | 0.6983 | 0.7040 | 0.7596 | 0.5433 |
| **SiamCRNN_GRU** | 0.9778 | 0.7021 | 0.7333 | 0.6949 | 0.7136 | 0.7660 | 0.5547 |
| **SiamCRNN_LSTM** | 0.9813 | 0.7287 | 0.8301 | 0.6648 | 0.7383 | 0.7830 | 0.5852 |
| **FCD-GAN (0.5)** | **0.9835** | **0.7671** | **0.8413** | **0.7197** | **0.7756** | **0.8083** | **0.6336** |

TABLE 2 ACCURACY ASSESSMENT ON HY DATASET

|  | OA | KC | Pre. | Rec. | F1 | mIOU | cIOU |
|---|---|---|---|---|---|---|---|
| **RNN** | 0.9438 | 0.7296 | 0.8051 | 0.7222 | 0.7614 | 0.7765 | 0.6147 |
| **DSCN** | 0.9340 | 0.6709 | 0.7871 | 0.6428 | 0.7077 | 0.7379 | 0.5476 |
| **SiamCRNN_FC** | 0.9454 | 0.7254 | 0.8500 | 0.6803 | 0.7557 | 0.7739 | 0.6074 |
| **SiamCRNN_GRU** | 0.9508 | 0.7533 | 0.8745 | 0.7051 | 0.7807 | 0.7932 | 0.6403 |
| **SiamCRNN_LSTM** | 0.9539 | 0.7701 | 0.8850 | 0.7228 | 0.7957 | 0.8051 | 0.6608 |
| **FCD-GAN (0.5)** | **0.9615** | **0.8190** | 0.8660 | **0.8177** | **0.8409** | **0.8413** | **0.7255** |

The optimizer used is Adam with warmup strategy. The initial epochs for generator training and segmentor training are both 50. The epochs for iteration optimization of generator and segmentor is 150. The batch size is 10. The probability threshold to determine change/non-change in the output of segmentor is fixed to be 0.5. The normalization to be zero mean and unit variance is applied as pre-processing. The weight of $\ell_1$ loss is 0.75 for WH dataset, and 0.65 for HY dataset. The content weight in generation loss is 0.2 for WH dataset, and 0.65 for HY dataset.

Table 1 and Table 2 show the accuracy assessments on WH and HY dataset. The comparative methods chosen is our previous work with pre-detection and deep learning network [44]. The result of FCD-GAN is the average of 5 time run. mIOU and cIOU are the values of the mean IOU and the only change IOU. OA indicates overall accuracy, KC indicates Kappa coefficient, Pre. indicates precision rate, and Rec. indicates recall rate. The bold values are the best accuracies, and the underline indicates the second best. It can be seen that, FCD-GAN achieved the highest accuracies in almost all indicators with an obvious increase.

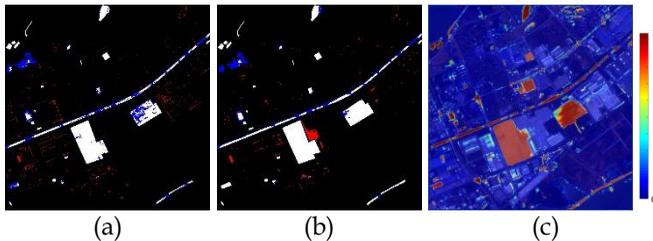

Figure 9 Change detection result of (a) SiamCRNN_LSTM and (b) FCD-GAN, where white indicates true detection, red indicates false alarms, and blue indicates omission error, as well as (c) the change probability output of FCD-GAN for HY dataset.

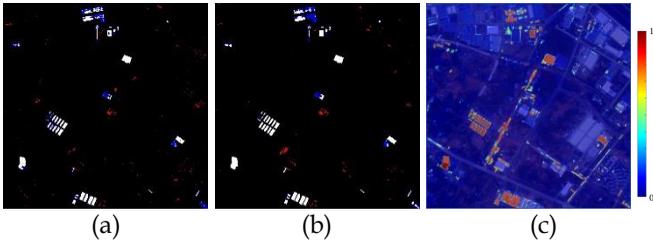

Figure 10 Change detection result of (a) SiamCRNN_LSTM and (b) FCD-GAN, where white indicates true detection, red indicates false alarms, and blue indicates omission error, as well as (c) the change probability output of FCD-GAN for WH dataset.

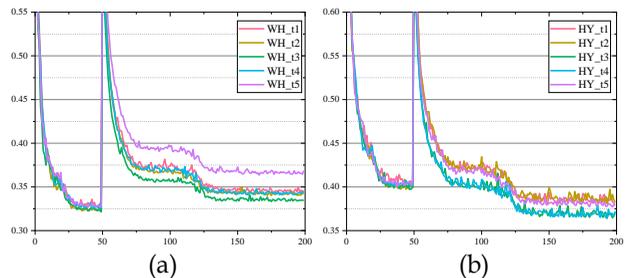

Figure 11 Generation loss of unsupervised change detection for (a) WH dataset and (b) HY dataset.

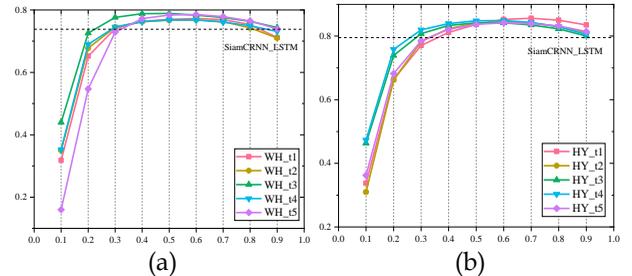

Figure 12 F1 score of FCD-GAN with different thresholds in (a) WH dataset and (b) HY dataset. The horizontal dash line indicates the accuracy of SiamCRNN_LSTM.

The generation loss for these two datasets is shown in Figure 11. 5-time runs are all shown in these figures. The losses before 50 is obtained with the whole image, and those afterwards is obtained with segmentor. It can be seen that the model converged after the iterations. Figure 12 shows the F1-score with different thresholds compared with the best method before - SiamCRNN_LSTM. For most thresholds around 0.5, the proposed FCD-GAN can get the best performance.

The assessments of binary change detection results are shown in Figure 9 and Figure 10. The references are used to draw the true detection, false alarm, and omission error with different colors. The 'jet' figures of change probability are also shown with an overlap on the original image. It can be observed that the proposed FCD-GAN find most of the changes and highlight the real changes obviously.

### 4.2 Weakly Supervised Change Detection

In the experiment of WSCD, there are two kinds of losses in FCD-GAN, which are generation loss and discrimination loss. If only the generation loss is used, the task of WSCD is changed to USCD. The generation loss is used to find the low-level changes, and the discrimination loss aims at find the high-level and semantic changes. Compared with USCD, which wants to find all landscape changes, WSCD only focus on the target changes, such as the building change in the dataset of BCD. So, the discrimination loss is the main loss.

The input image pairs are pre-processed with a global normalization to be zero mean and unit variance. In the optimization, generator is trained with the labelled unchanged image pairs, by the Adam optimizer, and frozen in the following optimization. The segmentor and discriminator are optimized by RMSProp optimizer. Since it is important to keep balance in the adversarial process, the learning rate of discriminator is assigned much lower than that of segmentor, due to the fact that the task of change discrimination is easier than change segmentation. The epochs for the training of generation and adversarial process are both 50. The batch size for the training of generation and adversarial process are 50 and 20, separately. Warm-up strategy is also used.

For generator, the weight of content loss kept 0.5 in training. In the learning of FCD-GAN only with generation loss (FCD-GN), which means we only use a USCD model, the weights of $\lambda_1$, $\lambda_2$ and $\lambda_3$ are 0.5, 1, and 1, while the discrimination loss is not used in training segmentor. In normal training, the weight of the main loss, which is the discrimination loss, always kept 1, like (8). In the learning of FCD-GAN only with adversarial process and without generator (FCD-AN), the weights of $\lambda_1$, $\lambda_2$, and $\lambda_3$ are 0.5, 1.5 and 0. In the final learning of FCD-GAN with all losses, the weights of $\lambda_1$, $\lambda_2$, and $\lambda_3$ are 1.6, 1.5 and 0.2.

There are 9935 unchanged pairs and 2616 changed pairs in BCD dataset. In order to take advantage of all sample pairs, in each epoch, the changed pairs are repeated and randomly ordered to fit the unchanged pairs.

Since there is no WSCD researches before, we selected the widely used CAM method in weakly supervised segmentation of computer vision [47]. It worth noting that the proposed FCD-GAN is only a basic framework, and we doesn't add some advanced modules and tricks, so the compared method only choose the basic CAM method [47]. The proposed FCD-GAN with/without generator or discriminator is also compared. Because in the evaluation of CAM, we also choose the threshold for the best performance, thus we use a common threshold 0.5 and the best threshold 0.9 for evaluation.

Table 3 shows the accuracy assessment on BCD dataset. Using CAM as the baseline, it can be observed that FCD-GAN with adversarial process can obtain a comparable and even better performance than the widely used CAM method. FCD-GN with only generation loss got a very low precision rate, since USCD finds all landscape changes, whereas many of them are not the target semantic changes. FCD-GAN only with adversarial process can get a similar accuracy with CAM, which means GAN structure can also be used in weakly supervised segmentation. FCD-GAN with a small weight of generation loss show the highest accuracy in Kappa, F1, mIOU, and cIOU.

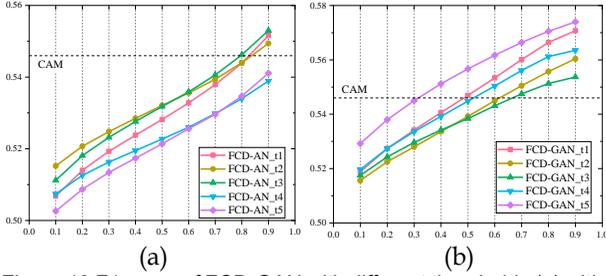

Figure 13 F1 score of FCD-GAN with different thresholds (a) without generator and (b) with generator. The horizontal dash line indicates the accuracy of CAM.

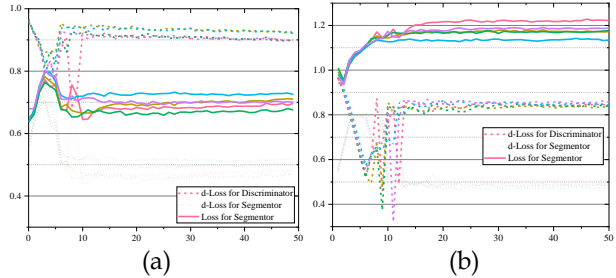

Figure 14 Loss of weakly supervised change detection (a) without generator and (b) with generator.

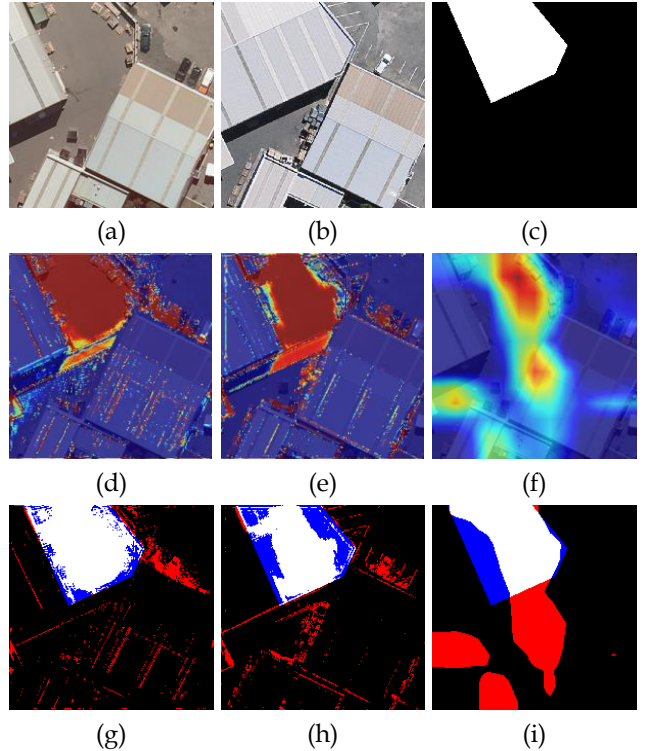

Figure 15 Example of WSCD in BCD dataset, which are (a) before change, (b) after change, (c) pixel-level reference, (d) density of FCD-AN, (e) density of FCD-GAN, (f) density of CAM, (g) binary change map of FCD-AN, (h) binary change map of FCD-GAN, and (i) binary change map of CAM, where white indicates true detection, red indicates false alarms, and blue indicates omission error.

TABLE 3 ACCURACY ASSESSMENT ON BCD DATASET

|  | OA | KC | Pre. | Rec. | F1 | mIOU | cIOU |
|---|---|---|---|---|---|---|---|
| **CAM** | **0.7929** | 0.4138 | **0.4940** | 0.6102 | 0.5460 | 0.5695 | 0.3755 |
| **FCD-GN (0.5)** | 0.5614 | 0.1667 | 0.2846 | 0.7594 | 0.4141 | 0.3710 | 0.2611 |
| **FCD-AN (0.5)** | 0.7055 | 0.3484 | 0.3921 | **0.8046** | 0.5272 | 0.5028 | 0.3580 |
| **FCD-GAN (0.5)** | 0.7526 | 0.3895 | 0.4364 | 0.7269 | 0.5452 | 0.5421 | 0.3748 |
| **FCD-GN (0.9)** | 0.7254 | 0.2655 | 0.3771 | 0.5302 | 0.4407 | 0.4874 | 0.2826 |
| **FCD-AN (0.9)** | 0.7429 | 0.3865 | 0.4270 | 0.7602 | 0.5468 | 0.5360 | 0.3763 |
| **FCD-GAN (0.9)** | 0.7898 | **0.4303** | 0.4893 | 0.6678 | **0.5645** | **0.5749** | **0.3932** |

In the learning of GAN structure, one important thing is whether the adversarial process is achieved and the discriminator cannot distinguish the fake sample accurately. The discrimination loss of FCD-AN and FCD-GAN is shown in Figure 14. It illustrates that the discrimination loss $\mathcal{L}_d^c$ for segmentor converged around 0.5, and the discrimination loss $1 - \mathcal{L}_d^c + \mathcal{L}_d^u$ for discriminator converged around 1. This phenomenon proved that it's hard for the discriminator to distinguish the real unchanged par and the mask changed pair. The adversarial process kept balanced.

The examples of segmentation densities and binary change maps are shown in Figure 15 and Figure 16. We can find some interesting phenomena. Compared with CAM, the proposed FCD-GAN can assign change density to more precise outline. Of course, CAM can refine this result with attention mechanism [48], post-process [49] etc., but we only focus on the most basic structure in this paper. The result from CAM is more integrated and regional, which is the direction of future development of FCD-GAN. The CAM only assign high density in the center of the changed landscapes, while FCD-GAN can label almost the whole shape of changed landscapes with the highest weight. CAM finds some false alarms, which are changed but not the target semantic building changes. FCD-GAN performs better in some aspects, whereas it highlighted the changed vehicles as well.

The experiments indicate that, WSCD is a very interesting topic. However, there are also lots problems in its performance. GAN structure can also be used in weakly supervised segmentation, which provides new potentials in this field.

### 4.3 Regional Supervised Change Detection

Compared with WSCD task, the semantic supervised information is provided by the given regions in RSCD task. What we can only know is that there are no changes outside the region. So, the model needs to find a minimum change mask, so that the remaining landscapes has the similar temporal pattern as those outside the given candidate regions.

The discrimination loss is used to satisfy the semantic supervision. The optimization process is similar with WSCD in previous sub-section. The generator is optimized with the whole training sets by masking the given region reference. Since the image pairs in OSCD dataset have different sizes, they were all sliced into 200 × 200 with an overlapping of 10 pixels, just like USCD experiment. The difference between implementation of RSCD and WSCD is that, the unchanged pairs are simulated with the image replacement inside the region reference. In OSCD dataset [46], 14 multi-temporal pairs are used for training, and 10 multi-temporal pairs are selected for testing.

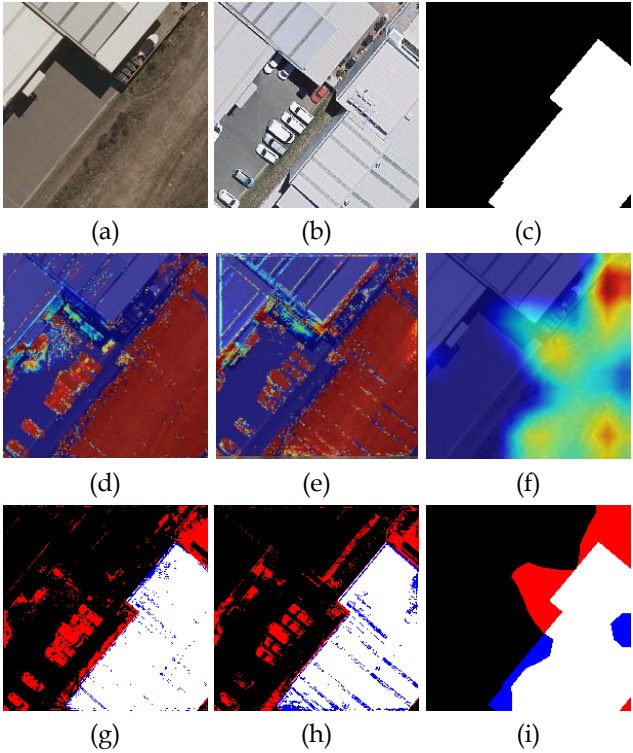

Figure 16 Example of WSCD in BCD dataset, which are (a) before change, (b) after change, (c) pixel-level reference, (d) density of FCD-AN, (e) density of FCD-GAN, (f) density of CAM, (g) binary change map of FCD-AN, (h) binary change map of FCD-GAN, and (i) binary change map of CAM, where white indicates true detection, red indicates false alarms, and blue indicates omission error.

For generator, the content weight is assigned as 0.1. In the optimization of FCD-GN with only generator, the weights of $\lambda_1$, $\lambda_2$ and $\lambda_3$ are 0.1, 2, and 1. FCD-GN doesn't take the discrimination loss into consideration, thus it is an unsupervised model. For FCD-AN with only discriminator and without generator, the weights of $\lambda_1$, $\lambda_2$ and $\lambda_3$ are 0.1, 2, and 0. FCD-AN is totally trained with the regional supervised information. If the discrimination and generation loss are both considered, FCD-GAN assign the weights of $\lambda_1$, $\lambda_2$ and $\lambda_3$ as 0.02, 2, and 0.5.

Since RSCD is a novel task and has no previous studies, we compare the proposed methods with the fully supervised model, which utilize the same segmentor and is supervised by a BCE loss to the pixel-level reference. Adam optimizer is applied with a 0.001 $\ell_2$ regularization.

Table 4 shows the accuracy assessment on OSCD dataset. Obviously, RSCD model cannot be comparable to FSCD model, whereas we can have a baseline for the performance of RSCD. Firstly, it can be found that the unsupervised model FCD-GN also got a good performance. It is because in the change detection with moderate-solution images, such as this OSCD dataset acquired by Sentinel-2 with the resolution of 10m, most labelled changes are interpreted according to spectral changes, which can be cap-tured by USCD model. Then, FCD-AN model without generator also shows satisfactory accuracies. It indicates that the proposed adversarial process is effective in interpreting the semantic supervised information. Finally, FCD-GAN with both discriminator and generator get the highest accuracy. When FSCD model can only get an accuracy of 0.4044, the F1-score of 0.3508 for FCD-GAN is good enough considering RSCD is only trained with such a few supervision information.

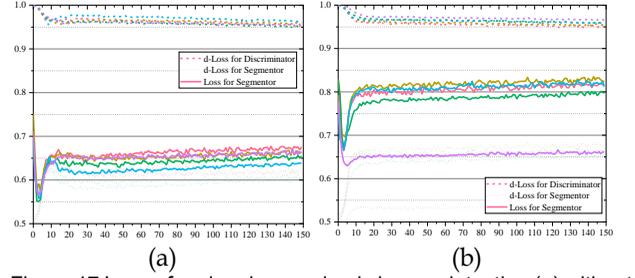

Figure 17 Loss of regional supervised change detection (a) without generator and (b) with generator.

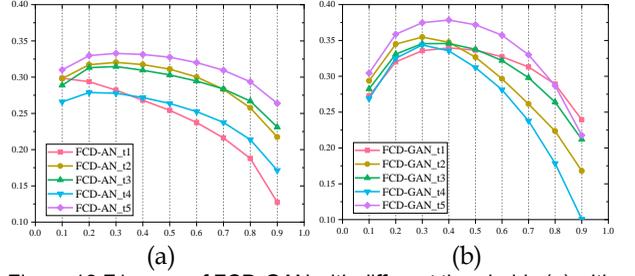

Figure 18 F1 score of FCD-GAN with different thresholds (a) without generator and (b) with generator.

TABLE 4 ACCURACY ASSESSMENT ON OSCD DATASET

| | OA | KC | Pre. | Rec. | F1 | mIOU | cIOU |
|---|---|---|---|---|---|---|---|
| **FSCD** | 0.9505 | 0.3801 | 0.5341 | 0.3254 | 0.4044 | 0.6015 | 0.2534 |
| **FCD-GN (0.5)** | 0.9306 | 0.2737 | 0.3192 | 0.3019 | 0.3102 | 0.5565 | 0.1836 |
| **FCD-AN (0.5)** | 0.9363 | 0.2594 | 0.3430 | 0.2557 | 0.2918 | 0.5533 | 0.1712 |
| **FCD-GAN (0.5)** | 0.9338 | 0.3022 | **0.3496** | 0.3259 | 0.3369 | **0.5677** | 0.2027 |
| **FCD-GN (0.3)** | 0.8978 | 0.2678 | 0.2425 | **0.4593** | 0.3173 | 0.5420 | 0.1886 |
| **FCD-AN (0.3)** | 0.9270 | 0.2671 | 0.3012 | 0.3124 | 0.3055 | 0.5531 | 0.1805 |
| **FCD-GAN (0.3)** | 0.9138 | **0.3071** | 0.2875 | 0.4504 | **0.3508** | 0.5623 | **0.2128** |

Figure 17 illustrates that the adversarial process kept balanced in training. And the segmentor is effective in confusing the discriminator. Figure 18 shows the F1-score in accordance with thresholds. The threshold of 0.3 can lead to the best performance in this experiment.

Figure 19 shows the change densities of FCD-AN, FCD-GAN and FSCD model in the testing pair covering Montpellier. It illustrates that the proposed FCD-GAN has the ability to find out most changes in this image. However, FCD-GAN tends to highlight more changes compared with the ground-truth. It could also be expected, since the region-level supervision can only point out some kinds of non-changes outside the given region. Considering the change types are extremely diverse, some types of changes, which is not the target in the dataset, cannot be excluded only with region-level references.

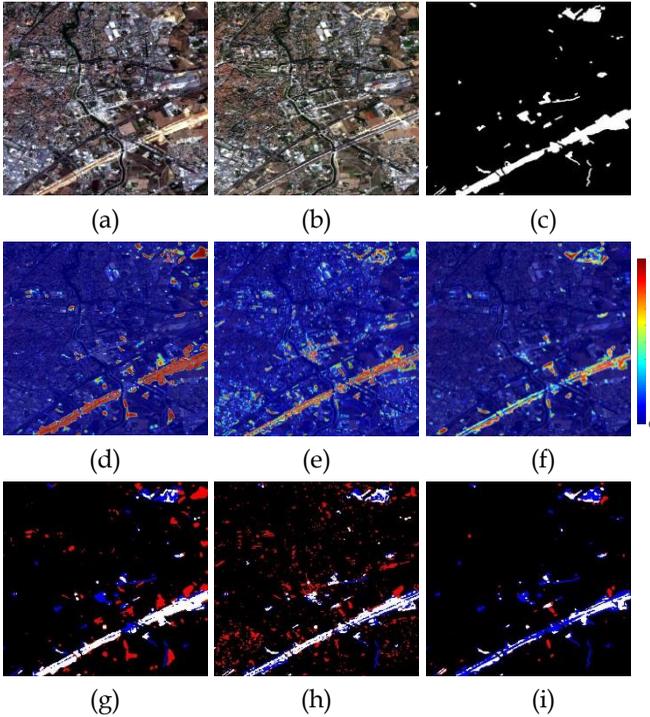

Figure 19 Example of RSCD in OSCD dataset, which covers Montpellier acquired on (a) May 15, 2016, (b) October 30, 2017, (c) pixel-level reference, (d) density of FCD-AN, (e) density of FCD-GAN, (f) density of FSCD, (g) binary change map of FCD-AN, (h) binary change map of FCD-GAN, and (i) binary change map of FSCD, where white indicates true detection, red indicates false alarms, and blue indicates omission error.

## 5 DISCUSSIONS

### 5.1 Difference from FSCD to RSCD and WSCD

Since change detection is concerned with multi-temporal images, it is more difficult to label all changed pixels for providing large-scale training dataset than other remote sensing tasks. So, as alternatives, RSCD and WSCD are valuable in practical applications for interpreters to provide a mass of training samples quickly and easily.

It is reasonable that the accuracies of RSCD and WSCD cannot be as high as FSCD, since they are not supervised with the same quantity of information. However, what is the exactly theoretical difference from FSCD to RSCD and WSCD?

As we analyze, the difference is changing from labelling target changes to labelling semantic non-changes.

For example, if there are 5 classes of landscapes in each image, the multi-temporal image pair will have $5 \times 5 = 25$ possible types of from-to changes, certainly 5 of them are unchanged. In FSCD, we may only focus on 4-6 types of changes, including vegetation-building, soil-building, building-soil, etc. Comparatively, RSCD and WSCD needs to provide the remaining 20 types of training samples in the unchanged pairs or outside the reference regions. This is because the basic assumption of RSCD and WSCD is to mask the changes so that the testing images will be similar with the unchanged ones in semantic aspects. If not, some unlabeled but untargeted false alarms in the final detection result is unavoidable.

So, theoretically, RSCD and WSCD needs abundant samples to reach the same performance of FSCD, although the samples in these two tasks are much easier to label.

### 5.2 Discussion about the constraints

In the framework, since we have separated the segmentor independently, the core of different change detection task is the corresponding constraints.

For unsupervised change detection, it is well known that the key for a good USCD algorithm is to build a better model for describing the temporal correlation between multi-temporal remote sensing images [50]. The temporal variations, that are anomalous from the correlations, can be distinguished as changes [16]. So, in the proposed framework, we utilize the powerful generator with spectral and content losses to build the relationship. In the basic format, we only train a unidirectional generator. The bidirectional generator is worth exploring in the future work.

For weakly and regional supervised change detection, since semantic supervision is added to define what is actually the changes, discriminator and adversarial process are utilized as the constraints to fuse the semantic information. Even though GAN structure has appeared in some previous change detection works, except for the translation between multi-temporal and multi-source data [23, 38, 51] or sample expansion [52], they are mostly used as a loss for fully supervised change detection [30, 32, 53, 54]. In our opinion, pixel-level loss is better for the segmentor in FSCD, and GAN is more suitable to the unequal supervision between the output of segmentor and reference. Discriminator can build the loss for the segmented changes and weakly/regional supervision, and passes the gradient to achieve the optimization; adversarial process can fine-tune the decision boundary.

For fully supervised change detection in the proposed framework, we also recommend the pixel-level loss as the constraints. The widely used BCE loss, DICE loss, focal loss etc. can be chosen. There are lots of works about the FSCD [31, 33, 34], thus it is not the focus of this paper. The reason why we do not recommend the discrimination loss is that, in the pooling process from the pixel-level to the image-level, the details of segmentation may be lost.

### 5.3 Limitation

Since the focus of this paper is to propose a novel framework of fully convolutional change detection with generative adversarial network, we didn't explore the detailed modules and parameter settings in the network. The segmentor, generator and discriminator are all very basic, to prove the effectiveness of the proposed framework.

According to current researches about change detection, many advanced modules can be added for better performance. For example, different fusion strategies can be considered in segmentor [28]; attention and transformer module can be used in network [33-35]; deep supervision and other tricks can also be implemented [29, 31]. For adversarial process, cycleGAN, bigGAN and other GAN network structures are feasible to be used [55, 56]. There is no doubt that with these advanced modules, better performance can be obtained in these datasets.

The main contribution of this paper is to propose a general framework with segmentor, generator and discriminator to fit all tasks of change detection. We propose how to utilize GAN in detecting semantic changes with only a few supervisions. The tasks of weakly supervised and regional supervised change detection are defined, and the experimental procedures are detailed introduced. We hope that, our research can provide new potentials for the field of change detection.

## 6 CONCLUSION

Change detection with deep learning networks has been one of the hottest topics in the field of remote sensing in recent years. For unsupervised change detection, networks are mostly used as feature extractor, or trained with pre-detection results. Only for fully supervised change detection task, the end-to-end segmentation networks are studied. However, it is hard to label abundant change detection datasets, since it is too labor-consuming with a more complex requirement.

Therefore, in this paper, we want to propose a general end-to-end framework of fully convolutional change detection with generative adversarial network (FCD-GAN), which can deal with all available change detection tasks. The change detection tasks are summarized as unsupervised change detection (USCD), weakly supervised change detection (WSCD), regional supervised change detection (RSCD), and fully supervised change detection (FSCD). Among them, WSCD and RSCD are newly defined to fit the requirements of real applications.

This framework contains three basic modules: segmentor, generator and discriminator. Segmentor and generator are combined with iterative optimization to solve USCD problem; segmentor and discriminator are trained with adversarial process to deal with WSCD and RSCD; segmentor with pixel-level loss can be used for FSCD, which has been intensively studied thus not the focus of this paper.

Three experimetns indicate the effectiveness of the proposed FCD-GAN in USCD, WSCD, and RSCD. FCD-GAN shows obvious improvement than the previous methods in USCD; it obtains a comparable result in WSCD with the classical CAM, which is the first attempt to take advantage of GAN in one-stage weakly segmentation; the proposed method also obtains a satisfactory performance in RSCD, which is a new and useful exploration in change detection.

Although the proposed framework is proved to be effective in the tasks of change detection, there are still some problems to be solved:

Firstly, the change detection result of the proposed FCD-GAN is precise in their outlines for WSCD. However, there are still some independent pixels inside the segmented regions. How to keep the completeness of the changed object is the first problem.

Secondly, for WSCD and RSCD, we generate the experimental data with the shared FSCD datasets. However, it can also be observed the accuracies are not very high. Therefore, other datasets, which are more suitable for the tasks of WSCD and RSCD, should be built to better evaluate the newly proposed algorithms, and other advanced modules can be added to build a useful model.

Finally, it is indicated that GAN can be used in weakly supervised segemention. We will test this structure in other computer vision dataset, and make meaningful exploration.

## ACKNOWLEDGEMENT

This work was supported in part by This work was supported in part by National Natural Science Foundation of China under Grant T2122014 and 61971317, and in part by the Natural Science Foundation of Hubei Province under Grant 2020CFB594, and in part by the Science and Technology Major Project of Hubei Province (Next-Generation AI Technologies) under Grant 2019AEA170. The corresponding authors are B. Du and L. Zhang.

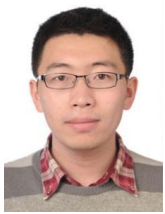

**Chen Wu** (M'16) received B.S. degree in surveying and mapping engineering from Southeast University, Nanjing, China, in 2010, and received the Ph.D. degree in Photogrammetry and Remote Sensing from State Key Lab of Information Engineering in Surveying, Mapping and Remote sensing, Wuhan University, Wuhan, China, in 2015. He is currently an Associate Professor with the State Key Laboratory of Information Engineering in Surveying, Mapping and Remote Sensing, Wuhan University, Wuhan, China. His research interests include multitemporal remote sensing image change detection and analysis in multispectral and hyperspectral images.

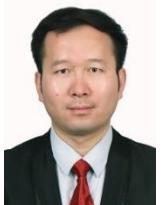

**Bo Du** (M'10–SM'15) received the Ph.D. degree in Photogrammetry and Remote Sensing from State Key Lab of Information Engineering in Surveying, Mapping and Remote Sensing, Wuhan University, Wuhan, China in 2010. He is currently a professor with the School of Computer Science and Institute of Artificial Intelligence, Wuhan University. He is also the director of and National Engineering Research Center for Multimedia Software Wuhan University, Wuhan, China. He has more than 80 research papers published in the IEEE Transactions on image processing (TIP), IEEE Transactions on cybernetics (TCYB), IEEE Transactions on Pattern Analysis and Machine Intelligence(TPAMI), IEEE Transactions on Geoscience and Remote Sensing (TGRS), etc. Fourteen of them are ESI hot papers or highly cited papers. His major research interests include machine leanring, computer vision, and image processing. He is currently a senior member of IEEE. He serves as associate editor for Neural Networks, Pattern Recognition and Neurocomputing. He also serves as a reviewer of 20 Science Citation Index (SCI) magazines including IEEE TPAMI, TCYB, TGRS, TIP, JSTARS, and GRSL. He won the Highly Cited Researcher (2019\2020) by the Web of Science Group. He won IEEE Geoscience and Remote Sensing Society 2020 Transactions Prize Paper Award. He won the IJCAI (International Joint Conferences on Artificial Intelligence) Distinguished Paper Prize, IEEE Data Fusion Contest Champion, and IEEE Workshop on Hyperspectral Image and Signal Processing Best paper Award, in 2018. He regularly serves as senior PC member of IJCAI and AAAI. He served as area chair for ICPR.

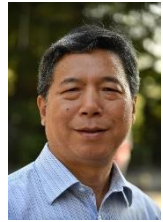

**Liangpei Zhang** (M'06–SM'08–F'19) received the B.S. degree in physics from Hunan Normal University, Changsha, China, in 1982, the M.S. degree in optics from the Xi'an Institute of Optics and Precision Mechanics, Chinese Academy of Sciences, Xi'an, China, in 1988, and the Ph.D. degree in photogrammetry and remote sensing from Wuhan University, Wuhan, China, in 1998. He is a chair professor in state key laboratory of information engineering in surveying, mapping, and remote sensing (LIESMARS), Wuhan University. He was a principal scientist for the China state key basic research project (2011–2016) appointed by the ministry of national science and technology of China to lead the remote sensing program in China. He has published more than 700 research papers and five books. He is the Institute for Scientific Information (ISI) highly cited author. He is the holder of 30 patents. His research interests include hyperspectral remote sensing, high-resolution remote sensing, image processing, and artificial intelligence. Dr. Zhang is a Fellow of Institute of Electrical and Electronic Engineers (IEEE) and the Institution of Engineering and Technology (IET). He was a recipient of the 2010 best paper Boeing award, the 2013 best paper ERDAS award from the American society of photogrammetry and remote sensing (ASPRS) and 2016 best paper theoretical innovation award from the international society for optics and photonics (SPIE). His research teams won the top three prizes of the IEEE GRSS 2014 Data Fusion Contest, and his students have been selected as the winners or finalists of the IEEE International Geoscience and Remote Sensing Symposium (IGARSS) student paper contest in recent years. He is the Founding Chair of the IEEE Geoscience and Remote Sensing Society (GRSS) Wuhan Chapter. He also serves as an Associate Editor or Editor for more than ten international journals. He serves as an Associate Editor for the IEEE TRANSACTIONS ON GEOSCIENCE AND REMOTE SENSING.